\documentclass[10pt, a4paper, conference, compsocconf]{IEEEtran}
\usepackage[utf8]{inputenc}
\usepackage{cite}
\usepackage{todonotes}
\usepackage[cmex10]{amsmath}
\usepackage{url}
\usepackage{graphicx}
\usepackage{lipsum}
\usepackage{listings}
\usepackage{bm}
\usepackage{hyperref}

\title{Improving OCR Accuracy on Early Printed Books\\by utilizing Cross Fold Training and Voting}
\author{\IEEEauthorblockN{Christian Reul$^1$, Uwe Springmann$^2$, Christoph Wick$^1$, and Frank Puppe$^1$}
\IEEEauthorblockA{
$^1$Chair for Artificial Intelligence and Applied Computer Science\\
$^2$Kallimachos Center for Digital Humanities\\
University of Würzburg, Germany\\
Email: \{firstname.lastname\}@uni-wuerzburg.de}
}
\date{November 2017}

\begin{document}

\maketitle

\begin{abstract}
In this paper we introduce a method that significantly reduces the character error rates for OCR text obtained from OCRopus models trained on early printed books. The method uses a combination of cross fold training and confidence based voting. After allocating the available ground truth in different subsets several training processes are performed, each resulting in a specific OCR model. The OCR text generated by these models then gets voted to determine the final output by taking the recognized characters, their alternatives, and the confidence values assigned to each character into consideration. Experiments on seven early printed books show that the proposed method outperforms the standard approach considerably by reducing the amount of errors by up to 50\% and more.
\end{abstract}


\section{Introduction}
Mainly due to the introduction of LSTM based recognition engines OCR on even the earliest printed books is not only possible but very precise \cite{springmannluedeling2017}\cite{Reul:2017:CSH:3078081.3078098}. However, contrary to modern prints or prints from the 19th century, character accuracy rates (CAR) above 98\% for older historical printings can usually only be achieved by training an individual model for a specific book. This is due to the high variability among different typefaces (glyph shapes) used for early printed books, especially from the 15th or 16th century. In order to train book specific models a certain amount of ground truth (GT) is required. In case of OCRopus the GT consists of line images and the corresponding transcription which are fed to the OCR engine during the training phase. Our goal is to improve the OCR accuracy with a given amount of GT by training different models and use voting to combine them. Most approaches use voting with outputs generated by different OCR engines like OCRopus\footnote{\url{https://github.com/tmbdev/ocropy}}, Tesseract\footnote{\url{https://github.com/tesseract-ocr}} or ABBYY Finereader\footnote{\url{https://www.abbyy.com/finereader/}. ABBYY also uses a voting mechanism internally to get the best recognition results, see \url{https://abbyy.technology/en:features:ocr:voting-api}}. Nonetheless, the leading commercial OCR engine ABBYY Finereader fails to produce usable output for early printings such as incunabula due to the lack of trained recognition models. While it is possible to train individual models for early printed books using Tesseract \cite{kirchner2016ocr}, the GT production and training process is considerably more time consuming and leads to less accurate results compared to using OCRopus. In the absence of alternative OCR engines to generate variance, we propose a cross fold training approach on a given GT pool for a single OCR engine (OCRopus) which leads to several models with different characteristics. To improve the voting result further we utilize the intrinsic confidence value produced by OCRopus. This enables the voting not only to take the top-1 output character into account but the top-n alternatives weighted by their confidences. In addition to the OCRopus engine we use the ISRI analytic tools \cite{rice1996isri} for alignment and voting. 

The rest of the paper is structured as follows: Chapter 2 introduces and discusses related work on OCR on early printed books and voting. The methods applied are described in detail in chapter 3. In chapter 4 the results achieved on seven early printed books are evaluated. These results are discussed in chapter 5 before chapter 6 concludes the paper.

\section{Related work}
An overview regarding topics concerning the improvement of OCR accuracy through combination is given in \cite{handley1998improving}. Apart from different methods to combine classifiers string alignment approaches are discussed.

The voting method to improve OCR results obtained from a variety of commercial OCR engines is introduced in \cite{rice1992report}. As early as 1996 Rice et al. \cite{rice1996isri} released a collection of command line scripts for the evaluation of OCR results called the ISRI analytic tools. The tools contain a voting procedure which first aligns several outputs using the Longest Common Substring (LCS) algorithm \cite{rice1994algorithm} and then applies a majority vote including heuristics to break ties. The tools were used to evaluate the results of various commercial OCR engines in several competitions on modern prints (see \cite{rice1996fifth}, e.g.). This voting procedure was able to improve the CAR of five engines on English business letters from between 90.10\% and 98.83\% to 99.15\%.

A different approach to achieve variance among OCR outputs was proposed by Lopresti and Zhou \cite{lopresti1997using}, who simply scanned each page three times and ran a single OCR engine on them. Their consensus sequence voting procedure led to a reduction of error rates between 20\% and 50\% on modern prints resulting in a CAR of up to 99.8\%.

Boschetti et al. \cite{boschetti2009improving} achieved an average absolute gain of 2.59\% compared to the best single engine (ABBYY, up to 97\% CAR) by combining the outputs of three different engines on ancient Greek editions from the 19th and 20th century. They applied a progressive alignment which starts with the two most similar sequences and extends the alignment by adding sequences. Then, the character selection is performed by a Naive Bayes classifier.

In \cite{lund2011progressive} Lund et al. trained maximum entropy models on synthetic data using voting and dictionary features. On a collection of typewritten documents from Word War II their method achieved 24.6\% relative improvement over the word error rate of the best of the five employed OCR engines.

Wemhoener et al. \cite{wemhoener2013creating} proposed an approach for aligning and combining different OCR outputs which can be applied to entire books and even different editions of the same book. First, a pivot is chosen among the outputs. Then, all other outputs are aligned pairwise with the pivot by first finding unique matching words in the text pairs to align them using a LCS algorithm. By repeating this procedure recursively two texts can be matched in an efficient way. Finally, all pairs are aligned along the pivot and a majority vote determines the final result.

Azawi et al. \cite{al2015combination} used weighted finite-state transducers based on edit rules to align the output of two different OCR engines. Neural LSTM networks trained on the aligned outputs are used to return a best voting. Since the network has used plenty of training data similar to the test set, it is able to predict correct characters even in cases, where both engines failed. During tests on printings with German Fraktur and the UW-III data set\footnote{\url{http://isis-data.science.uva.nl/events/dlia//datasets/uwash3.html}} the LSTM approach led to CERs (character error rates) around 0.40\%, while the ISRI voting tool and the method presented in \cite{wemhoener2013creating} achieved between 1.26\% and 2.31\%. A principal drawback of this method is its reliance on fixed input-output relationships, i. e. each OCR token is mapped to a single 'correct' token. But historical spelling patterns are much more variable than modern ones and the same word is often spelled and printed in more than one form even in the same document. This method therefore not only corrects OCR errors but also normalizes historical spellings which may not be desired.

Our method shows considerable differences compared to the work presented above. Not only is it applicable to some of the earliest printed books, but it also works with only a single open source OCR engine. Furthermore, it can be easily adapted to practically any given book using a reasonable amount of GT without the need for excessive data to train on.

\section{Methods}
\label{sec:methods}
The general idea of the proposed approach is to significantly improve the accuracy that can be achieved by using only a single OCR engine. The trade-off is between adding more GT to the training pool (a costly manual process) and a considerable increase in the required computational effort. Here we take the second route with a given amount of GT keeping the additional manual effort to a minimum. In the following, we first introduce the workflow before describing the confidence based voting and the needed adaptations in the OCRopus engine in detail. The standard majority voting without confidence can be carried out by using OCRopus and the ISRI tools in their default configuration.

\subsection{Workflow}
The general workflow can be described as follows:\\
\textbf{Input:} Line-based GT consisting of line image and the corresponding transcription.\\
\textbf{Output:} Recognized text lines.
\begin{enumerate}
\item Divide the available GT in \textit{N} distinct folds and set aside some held out data for evaluation.
\item Train \textit{N} OCRopus models.
    \begin{enumerate}
    \item Declare one of the folds as test data and allocate the rest for training.
    \item Run a training using \textit{N}-1 folds as training data.
    \item Choose the best model by testing on the remaining fold.
    \end{enumerate}
\item Apply the \textit{N} trained models to previously unseen lines (the held out evaluation data) and determine the result by voting.
\end{enumerate}

\subsection{GT Allocation and Model Training}
For example, if the number of folds \textit{N} is 5 and there are 150 lines of GT available, each fold contains 30 lines. For the first training fold 1 is used for testing and folds 2 through 5 for training. For the second training, fold 2 for testing and folds 1, 3, 4, 5 for training, and so on. So in this example each OCRopus training is carried out on 120 lines, which represent 80\% of the GT pool. The entire training process closely follows the approach described in \cite{uwe_springmann_2016_46571}. After the training process is finished, the best of the resulting models for each fold is determined by recognizing the test lines with each model and select the one with the lowest CER. These five best models are then used to recognize the unseen lines of the held-out data resulting in five text outputs for each line, which then serve as input for the voting tool.

\subsection{Alignment}
As an example we use the text line from Figure \ref{fig:exampleLine}. The corresponding GT and the recognition results of five best models M1-M5 look like this:

\begin{figure}[t]
\centering
\includegraphics[width=0.9\linewidth]{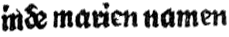}
\caption{Example text line containing a strongly degraded \textit{e}.}
\label{fig:exampleLine}
\end{figure}

\begin{lstlisting}
GT: inde marien namen
----------------------
M1: inide maricn namen
M2: inde maricn namen
M3: inde marien namen
M4: iade marien namen
M5: inde maricn namen
\end{lstlisting}

The alignment tool produces the following output using M1-M5 as inputs:
\begin{lstlisting}
Aligned: i{1}de mari{2}n namen
---------------------------------------
{1}: M1{ni}, M2{n}, M3{n}, M4{a}, M5{n}
{2}: M1{c}, M2{c}, M3{e}, M4{e}, M5{c}
\end{lstlisting}

The first line shows the aligned output. Curled parenthesis mark disagreements between two or more inputs. Afterwards, the different recognition results of the models are listed for each disagreement.

\subsection{Additional Information about the Recognition Confidence Values}
During the recognition process the OCRopus network determines the probabilities (represented by confidence values) of the output characters at each position in the text line as a distribution over the complete character set (the 'codec').
The size of this set depends on the individual model and gets fixed at the start of the
training process. 
To access this additional information some changes within the OCRopus code had to be made. The confidence values are collected and stored for each line in an so-called \textit{extended lloc} (\textbf{L}STM \textbf{l}ocation \textbf{o}f \textbf{c}haracters) file. See Table \ref{tab:llocs} for the first few \textit{llocs} of M4.

\begin{table}[t]
\renewcommand{\arraystretch}{1.3}
\centering
\caption{Example llocs output from M4 for the word 'inde' recognized as 'iade' including the most likely character, its start/end position, confidence and its alternatives including their respective confidences.}

\label{tab:llocs}
\begin{tabular}{ccccl}
\hline
\textbf{Char} & $\bm{x_S}$ & $\bm{x_E}$ & \textbf{Conf} & \multicolumn{1}{c}{\textbf{Alternatives}}                                                      \\
\hline
i                  & 120         & 123         & 87.54\%         & b=8.66\%, f=2.94\%                                                                   \\
\hline
a                  & 126         & 136         & 96.65\%       & \begin{tabular}[c]{@{}l@{}}n=45.78\%, r= 23.65\%,\\ m=9.24\%, k=8.32\%, {[}...{]}\end{tabular} \\
\hline
d                  & 142         & 149         & 99.93\%       & ã=4.83\%, V=4.17\%, O=1.13\%                                                                   \\
\hline
e                  & 155         & 160         & 99.15\%       &    all alternatives $<$ 1\%                \\
\hline
\end{tabular}
\end{table}

From left to right the columns show the most likely character, the pixel coordinates of its start and end position, and its confidence. The rightmost column contains a list of alternatives with their respective confidences. 


The sum of all confidence values (representing a posterior probability distribution over all output nodes, i.e. all possible characters) per pixel position adds up to 100\%. Because each glyph representing a character is several pixels wide, an alternative n-best recognition might occur at a different pixel position and the combined confidences of all alternatives will in general add up to values above 100\%.
E.g. see the second row in Table \ref{tab:llocs}: The recognized \textit{a} has a confidence maximum of 96.65\% somewhere between the start and end positions at 126 and 136 pixels, leaving only 3.35\% to be distributed among all remaining characters.
However, it is quite common that an alternative (in this example the \textit{n}) is recognized at a different position than the top-1 character (\textit{a})
with a confidence lower than the maximum (96.65\%) but significantly higher than the left over percentages at the position of the maximum. Obviously, the higher confidence at a different position is much more relevant to a confidence based voting.

\subsection{Confidence Voting}
After the alignment the confidence voting takes place. The aligned output is processed from left to right. Characters which could be matched for all inputs are accepted right away. To solve the disagreements between two or more inputs the corresponding \textit{llocs} are identified and loaded. Since the disagreements can vary in length, first a majority vote takes place to determine the most likely length. Longer or shorter inputs are discarded (e.g. the output of M1 \textit{ni} in the first disagreement). Of course, these inputs could still hold some valuable information, and therefore an alignment of these disagreements might make sense. But, preliminary tests showed that aligning the inputs varying in length by applying a k-means clustering on the recognition position led to a decline in accuracy for the voting result. There are two likely reasons for this: First, as explained above, the recognition position for a character can vary considerably along the glyph in the input image, and therefore lead to mismatches. Second, if a model already confused a single character as two or vice versa, the output information on the recognition output and the alternatives could be heavily flawed. For the unlikely case of a tie during the length voting, for example if the outputs have the lengths 1, 2, 2, 3, 3, the shorter option is chosen to break the tie. This heuristic was implemented as it counters a common OCRopus problem where the network recognizes the same character twice because an $\epsilon$ (no character) is recognized within a glyph, and consequently enables the network to perform another output. "correct"ing the inputs the confidences for the recognized characters and all relevant alternatives (confidence $>$ 1\%) are summed up and the most likely ones are accepted. 

In our example (see Table \ref{tab:confVote}) the heavily degraded \textit{e} at disagreement position 2 got wrongfully recognized as a \textit{c} by three out of the five models. Therefore, the simple majority vote leads to a \textit{c} in the final output as does the confidence voting when only considering the actually recognized character. However, when incorporating the alternatives the correct solution \textit{e} is chosen.

\begin{table}[t]
\renewcommand{\arraystretch}{1.3}
\centering
\caption{Confidences of the model outputs for the two characters in question (\textit{c} and \textit{e}) including the \underline{most likely character} and its alternative as well as the confidence sum without (\textit{Rec}) and with (\textit{+ Alt}) including the alternatives.}
\label{tab:confVote}
\begin{tabular}{ccc}
\hline
\textbf{}        & \textbf{c}        & \textbf{e}        \\
\hline
\textbf{M1}      & \underline{66.83\%}  & 38.40\%           \\
\textbf{M2}      & \underline{93.27\%}  & 19.77\%           \\
\textbf{M3}      & -                 & \underline{99.91\%}  \\
\textbf{M4}      & 7.56\%            & \underline{98.02\%}  \\
\textbf{M5}      & \underline{90.31\%}  & 50.07\%           \\
\hline
\textbf{$\sum$ Rec} & \textbf{250.41\%} & 197.93\%          \\
\textbf{$+ \sum$ Alt}   & 257.97\%          & \textbf{306.17\%} \\
\hline
\end{tabular}
\end{table}

\section{Experiments}
In this chapter we briefly introduce the data before describing the experiments and reporting the obtained results. Each model training was carried out until the recognition accuracy on the test set just showed variation due to statistical noise and no further improvement was expected.

\subsection{Data}
\begin{figure}[t]
\centering
\includegraphics[width=0.75\linewidth]{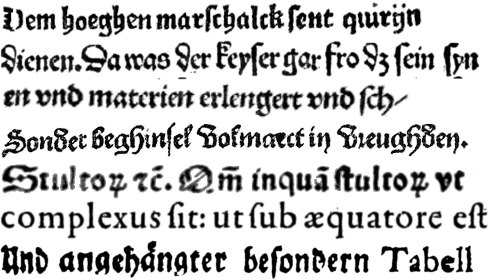}
\caption{Seven different example lines from the seven used books. From top to bottom: 1476, 1488, 1495, 1500, 1505, 1572, 1675.}
\label{fig:lines}
\end{figure}

The experiments were performed on seven early printed books (see Table \ref{tab:books}). To avoid unwanted side effects only lines from running text parts were used and headings, marginalia, page numbers etc. were excluded. Fig. \ref{fig:lines} shows some example lines.

\begin{table}[t]
\renewcommand{\arraystretch}{1.3}
\centering
\caption{Books used for Evaluation.}
\label{tab:books}
\begin{tabular}{cccc}
\hline
\textbf{ID/Year} & \textbf{Language} & \textbf{GT Train} & \textbf{GT Test} \\
\hline
1476             & German             & 1000              & 2000             \\
1488             & German             & 1000              & 3178             \\
1495             & German             & 1000              & 1114                         \\
1500             & Dutch              & 1000              & 1500             \\
1505             & Latin              & 1000              & 2289 \\
1572             & Latin              & 1000              & 541 \\
1675             & German             & 250               & 317             \\
\hline
\end{tabular}
\end{table}
1495, 1500 and 1505 are editions of the \textit{Ship of Fools} (\textit{Narrenschiff} by Sebastian Brant) and were digitized as part of an effort to support the \textit{Narragonien digital} project at the University of Würzburg\footnote{\url{http://kallimachos.de/kallimachos/index.php/Narragonien}}. Despite their almost identical content these books differ considerably from an OCR point of view since they were printed in different print shops using different typefaces and languages (Latin, German, and Dutch). 1488 was gathered during a case study of highly automated layout analysis \cite{Reul:2017:CSH:3078081.3078098}. 1476 is part of the Early New High German Reference Corpus\footnote{\url{http://www.ruhr-uni-bochum.de/wegera/ref/index.htm}} and 1572 was digitized for the Arabic-Latin AL-Corpus\footnote{\url{http://arabic-latin-corpus.philosophie.uni-wuerzburg.de/}}. Finally, 1675 was chosen from the RIDGES corpus\footnote{\url{http://korpling.org/ridges/}} \cite{springmannluedeling2017} for its high error rate. 

\subsection{Default Application (5 folds, 150 lines)}
As a first experiment the number of folds \textit{N} was set to 5 and 150 lines were used since this represents a magnitude which usually already yields good results without hitting the point of diminishing return. Moreover, most modern PCs with multiple cores should be able to comfortably handle five parallel OCRopus training processes. The experiment was conducted by following the workflow and example described in section \ref{sec:methods}. Table \ref{tab:5x150} shows the CER achieved on a fixed evaluation set of previously unseen lines from the held-out data of each individual best model (\textit{1-5}) and the combined results without (\textit{ISRI Voting}) and with (\textit{Confidence Voting}) confidence information. Furthermore, the relative improvement of the combined result with respect to the \textit{best/average/worst} model is indicated.

\begin{table*}[t]
\renewcommand{\arraystretch}{1.3}
\centering
\caption{CERs and improvement rates of (confidence) voting over the best, average and worst result of single models when using 5 folds and 150 lines.}
\label{tab:5x150}
\begin{tabular}{c|ccccc|cccc|cccc}
\hline
\textbf{}                         & \multicolumn{5}{c|}{\textbf{CERs of the Best Model of each Individual Fold}}                                                                                      & \multicolumn{4}{c|}{\textbf{ISRI Voting}}                                                                                          & \multicolumn{4}{c}{\textbf{Confidence Voting}}                                                                                                  \\
\hline
                              \textbf{Year}     & \textbf{1}                 & \textbf{2}                 & \textbf{3}                 & \textbf{4}                 & \textbf{5}                 & \textbf{CER}               & \multicolumn{3}{c|}{\textbf{\begin{tabular}[c]{@{}c@{}}Improvement over\\ best/avg/worst\end{tabular}}} & \textbf{CER}                        & \multicolumn{3}{c}{\textbf{\begin{tabular}[c]{@{}c@{}}Improvement over\\ best/avg/worst\end{tabular}}}      \\
\hline
\textbf{1476}                     & 3.93\%                     & 3.32\%                     & 4.07\%                     & 3.61\%                     & 3.41\%                     & 2.21\%                     & 35\%                             & 38\%                            & 44\%                            & \textbf{1.82\%}                     & \textbf{45\%}                     & \textbf{50\%}                     & \textbf{55\%}                     \\
\textbf{1488}                     & 2.87\%                     & 2.58\%                     & 2.54\%                     & 2.34\%                     & 4.23\%                     & 1.60\%                     & 32\%                             & 45\%                            & 62\%                            & \textbf{1.42\%}                     & \textbf{40\%}                     & \textbf{51\%}                     & \textbf{67\%}                     \\
\textbf{1495}                     & 3.97\%                     & 5.21\%                     & 6.16\%                     & 6.34\%                     & 4.34\%                     & 3.52\%                     & 11\%                             & 32\%                            & 44\%                            & \textbf{2.89\%}                     & \textbf{27\%}                     & \textbf{44\%}                     & \textbf{54\%}                     \\
\textbf{1500}                     & 3.10\%                     & 2.66\%                     & 2.87\%                     & 2.69\%                     & 2.82\%                     & 1.74\%                     & 35\%                             & 38\%                            & 44\%                            & \textbf{1.54\%}                     & \textbf{42\%}                     & \textbf{45\%}                     & \textbf{50\%}   \\
\textbf{1505}                     & 5.29\%                     & 5.21\%                     & 4.93\%                     & 5.57\%                     & 4.67\%                     & 3.96\%                     & 15\%                             & 24\%                            & 29\%                            & \textbf{3.70\%}                     & \textbf{21\%}                     & \textbf{29\%}                     & \textbf{34\%}                     \\
\textbf{1572}                     & 1.62\%                     & 1.95\%                     & 1.74\%                     & 2.02\%                     & 1.99\%                     & 1.49\%                     & 8\%                             & 20\%                            & 26\%                            & \textbf{1.38\%}                     & \textbf{15\%}                     & \textbf{26\%}                     & \textbf{31\%}                     \\
\textbf{1675}                     & 10.93\%                    & 11.19\%                    & 11.81\%                    & 12.69\%                    & 11.26\%                    & 9.22\%                     & 16\%                             & 20\%                            & 27\%                            & \textbf{8.80\%}                     & \textbf{20\%}                     & \textbf{24\%}                     & \textbf{31\%}  \\
\hline
\end{tabular}
\end{table*}

The results clearly show that the cross fold training and voting process reduced the CER considerably for all books. The amount of errors corrected by combining the outputs varies from 16\% for the best individual models to 62\% for the worst ones. All improvements are highly significant on a better than 0.001 level with the $\chi^2$ test. Incorporating the confidence information approximately reduces the amount of errors by another 5\% to 10\%. This additional improvement is also highly significant except for 1572 and 1675 due to their relatively small number of lines used for the evaluation (541 and 317 respectively).

\subsection{Influence of the Number of Lines}
In the next experiment the influence of the number of lines on voting was studied by varying them in six steps between 60 and 1,000 (see Table \ref{tab:lines}). Because of the previous results only confidence voting is considered. Furthermore, since in a real world scenario lacking held-out GT data there is no way to determine the best or worst of the five individual models, only average improvement is noted.

\begin{table}[t]
\renewcommand{\arraystretch}{1.3}
\centering
\caption{CERs (top) and improvement rates (bottom) of confidence voting over the average result of single models when varying the amount of lines.}
\label{tab:lines}
\begin{tabular}{lcccccc}
\hline
              & \textbf{60}                                            & \textbf{100}                                          & \textbf{150}                                          & \textbf{250}                                          & \textbf{500}                                          & \textbf{1000}                                         \\
              \hline
\textbf{1476} & \begin{tabular}[c]{@{}c@{}}7.55\%\\ 41\%\end{tabular}  & \begin{tabular}[c]{@{}c@{}}2.66\%\\ 50\%\end{tabular} & \begin{tabular}[c]{@{}c@{}}1.82\%\\ 50\%\end{tabular} & \begin{tabular}[c]{@{}c@{}}1.46\%\\ 44\%\end{tabular} & \begin{tabular}[c]{@{}c@{}}1.20\%\\ 42\%\end{tabular} & \begin{tabular}[c]{@{}c@{}}0.93\%\\ 35\%\end{tabular} \\

\hline
\textbf{1488} & \begin{tabular}[c]{@{}c@{}}4.76\%\\ 32\%\end{tabular}  &    \begin{tabular}[c]{@{}c@{}}2.07\%\\ 53\%\end{tabular}                                                   & \begin{tabular}[c]{@{}c@{}}1.42\%\\ 51\%\end{tabular} & \begin{tabular}[c]{@{}c@{}}0.90\%\\ 45\%\end{tabular} & \begin{tabular}[c]{@{}c@{}}0.74\%\\ 39\%\end{tabular} & \begin{tabular}[c]{@{}c@{}}0.65\%\\ 34\%\end{tabular} \\
\hline
\textbf{1495} & \begin{tabular}[c]{@{}c@{}}8.05\%\\ 37\%\end{tabular}  &      \begin{tabular}[c]{@{}c@{}}4.01\%\\ 47\%\end{tabular}                                                 & \begin{tabular}[c]{@{}c@{}}2.89\%\\ 44\%\end{tabular} & \begin{tabular}[c]{@{}c@{}}1.99\%\\ 42\%\end{tabular} & \begin{tabular}[c]{@{}c@{}}1.57\%\\ 39\%\end{tabular} & \begin{tabular}[c]{@{}c@{}}1.34\%\\ 37\%\end{tabular} \\
\hline
\textbf{1500} & \begin{tabular}[c]{@{}c@{}}2.81\%\\ 41\%\end{tabular}  &       \begin{tabular}[c]{@{}c@{}}1.87\%\\ 43\%\end{tabular}                                                & \begin{tabular}[c]{@{}c@{}}1.54\%\\ 45\%\end{tabular} & \begin{tabular}[c]{@{}c@{}}1.23\%\\ 39\%\end{tabular} & \begin{tabular}[c]{@{}c@{}}1.06\%\\ 35\%\end{tabular} & \begin{tabular}[c]{@{}c@{}}0.97\%\\ 33\%\end{tabular} \\
\hline
\textbf{1505}  & \begin{tabular}[c]{@{}c@{}}5.82\%\\ 24\%\end{tabular}  &       \begin{tabular}[c]{@{}c@{}}4.24\%\\ 27\%\end{tabular}                                                & \begin{tabular}[c]{@{}c@{}}3.70\%\\ 29\%\end{tabular} & \begin{tabular}[c]{@{}c@{}}3.49\%\\ 28\%\end{tabular} & \begin{tabular}[c]{@{}c@{}}2.63\%\\ 29\%\end{tabular} & \begin{tabular}[c]{@{}c@{}}2.46\%\\ 27\%\end{tabular} \\
\hline
\textbf{1572}  & \begin{tabular}[c]{@{}c@{}}1.90\%\\ 35\%\end{tabular}  &       \begin{tabular}[c]{@{}c@{}}1.49\%\\ 29\%\end{tabular}                                                & \begin{tabular}[c]{@{}c@{}}1.38\%\\ 26\%\end{tabular} & \begin{tabular}[c]{@{}c@{}}1.22\%\\ 22\%\end{tabular} & \begin{tabular}[c]{@{}c@{}}0.98\%\\ 24\%\end{tabular} & \begin{tabular}[c]{@{}c@{}}0.73\%\\ 31\%\end{tabular} \\
\hline
\textbf{1675} & \begin{tabular}[c]{@{}c@{}}14.48\%\\ 19\%\end{tabular} &         \begin{tabular}[c]{@{}c@{}}10.03\%\\ 27\%\end{tabular}                                              & \begin{tabular}[c]{@{}c@{}}8.80\%\\ 24\%\end{tabular} & \begin{tabular}[c]{@{}c@{}}5.77\%\\ 32\%\end{tabular} & -                                                     & - \\
\hline
\end{tabular}
\end{table}

Cross fold training and confidence voting on 60 to 1,000 lines shows similar improvements as the previous experiments using 150 lines. Good outputs benefit more from voting than worse ones. However, the by far worst recognition result (1675, 60 lines) still shows a considerable improvement. For most books a medium amount of GT (150 to 250 lines) leads to the biggest decreases of the CER compared to the average of the individually trained models. It is especially noteworthy that voting is still very effective when combining the output of very high performance models: On all three books that surpassed a character accuracy rate of 99\% the average amount of remaining errors was reduced by at least one third.

\subsection{Influence of the Number of Folds}
To increase the degree of variety among the models even more the number of folds was set to 10. The size of the testing sets is given by the number of lines divided by the number folds. Consequently, a higher number of folds leads to less lines being used for testing and more for training. For example, when training five folds using 250 lines the train/test ratio is 200/50 for each model training. This ratio rises to 225/25 when increasing the number of folds to 10. Therefore, another experiment (5+) using five folds was conducted where fewer lines were added to the test sets in order to match the number of training lines during training with ten folds without altering the overall amount of GT lines. For practical reasons and since all books clearly showed the same tendencies, this final experiment was conducted using a subset of two books with comprehensive GT and varying CER: 1476 and 1505. Table \ref{tab:folds} shows the results.

\begin{table}[t]
\renewcommand{\arraystretch}{1.3}
\centering
\caption{CERs when using 5 or 10 folds with different size of the training set (5+).}
\label{tab:folds}
\begin{tabular}{c|ccc|ccc}
\hline
              & \multicolumn{3}{c|}{\textbf{1476}}      & \multicolumn{3}{c}{\textbf{1505}}      \\
              & \textbf{5} & \textbf{5+} & \textbf{10} & \textbf{5} & \textbf{5+} & \textbf{10} \\
\hline              
\textbf{150}  & 1.82\%     & 1.98\%      & 1.78\%      & 3.70\%     & 3.71\%      & 3.51\%      \\
\textbf{250}  & 1.46\%     & 1.57\%      & 1.24\%      & 3.49\%     & 3.31\%      & 3.15\%      \\
\textbf{1000} & 0.93\%     & 0.94\%      & 0.86\%      & 2.46\%     & 2.34\%      & 2.16\%     \\
\hline
\end{tabular}
\end{table}

Doubling the number of folds led to a decrease of the CER in all scenarios. This effect was smaller when using 150 lines. Adjusting the number of lines when using five folds always led to worse results compared to ten folds.

\subsection{Time Expenditure}
Table \ref{tab:time} shows the necessary computational time expenditure for different scenarios with a varying number of folds and a maximum number of training steps depending on the number of lines. All measurements were performed on a laptop with a quad core i5-6300HQ CPU @ 2.3 GHz and 8 GB RAM using multi-threading whenever possible. The speed of the OCRopus training and prediction process depends on the length of the lines. Therefore, these measurements were performed on the book 1500, since its line lengths are closest to the average of all used books. The time expenditure for the training setup, alignment, and voting is negligible. The results show that the benefits of reduced error rates of our method can be reached by doing the necessary model training overnight.

\begin{table}[t]
\renewcommand{\arraystretch}{1.3}
\centering
\caption{Required time expenditure for the processing of book 1500 (4651 lines) using different scenarios in terms of training Fold x Lines and number of training iterations.}
\label{tab:time}
\begin{tabular}{lcccc}
\hline
\multicolumn{1}{c}{}            & \textbf{\begin{tabular}[c]{@{}c@{}}5x150\\ 10k It.\end{tabular}} & \textbf{\begin{tabular}[c]{@{}c@{}}5x250\\ 20k It.\end{tabular}} & \textbf{\begin{tabular}[c]{@{}c@{}}5x1000\\ 30k It.\end{tabular}} & \textbf{\begin{tabular}[c]{@{}c@{}}10x1000\\ 30k It.\end{tabular}} \\
\hline
\textbf{Training}               & 177min                                                           & 238min                                                           & 381min                                                            & 782min                                                             \\
\textbf{Recognize Book} & 26min                                                             & 26min                                                             & 26min                                                              & 52min                                                              \\
\hline
\textbf{Sum}                    & \textbf{203min}                                                  & \textbf{264min}                                                  & \textbf{407min}                                                   & \textbf{834min} \\
\hline
\end{tabular}
\end{table}

\section{Discussion}
Our experiments show that the proposed approach significantly improves the obtainable character accuracy rate on early printed books. As expected, OCR texts with a lower CER gain an even bigger boost compared to the more erroneous results. However, while not reaching the same improvement rates of up to over 50\% even the worst OCR texts still benefitted greatly from gains of close to 30\%.

The number of available GT lines did not show a direct influence on the achievable results. Nonetheless, a very high number of lines leads to a drop in gains for most books. This has to be expected for models that get closer to perfection as most of the remaining errors are unavoidable ones such as missing characters or highly degraded glyphs leading to misrecognition by any model. Both of these cases cannot be restored by voting.

Our first experiments on increasing the number of folds showed promising signs, especially when training with a large pool of GT. Adjusting the train/test ratio towards more training lines led to varying results depending on the number of GT lines. For small to medium amounts the CER stayed the same or even went up compared to the standard 80\%/20\% ratio. This indicates that choosing the best model based on a small number of lines can lead to models that perform well on the test data, but don't generalize well.

Moreover, our approach allows for a considerably more efficient use of the available GT. For example, in almost all cases the cross fold training and confidence voting based on 100 lines of GT considerably outperformed the default single model approach even when using 50\% more lines.

Furthermore, it is worth mentioning that the CERs of the individually trained models vary strongly, producing up to 80\% more errors. This kind of variance represents a big problem for the standard approach. Obviously, in a real world scenario there is no way to determine if the training of a single model went well in terms of variance. Our robust approach doesn't suffer noticeably from a single flawed model, as is shown by e.g. 1488 in Table \ref{tab:5x150}.

Finally, the experiment regarding time expenditure showed that even very comprehensive multi-fold training tasks can be performed by a standard system within a reasonable amount of time, e.g. overnight.

\section{Conclusion and Future Work}
A method to significantly improve the CER on early printed books by utilizing cross fold training and confidence based voting was proposed. The results showed that our method works with any reasonable amount of GT, and is therefore applicable to all stages during the OCR process on an early printed book.

Despite the very encouraging results there is still some work to be done: First, the alignment process of the \textit{llocs} prior to confidence voting should be further optimized in order to allow an improved matching of disagreements with varying lenghts. Furthermore, there are several parameters to be optimized during the cross fold training process like the optimal number of folds and the train/test split within a fold. The best choice mainly depends on the amount of available GT, but can also vary because of the accessible hardware or the overall OCR quality of the individual book. To be able to provide reliable recommendations for different scenarios further extensive tests are required.

Another promising option is to benefit from the diversity of the models obtained during cross fold training even further by implementing an active learning approach. If additional training is required, the resulting outputs allow an informed instead of random selection of new GT lines, e.g. by choosing the lines whose outputs differ the most, indicating a high degree of uncertainty across the models.

Despite the focus on early printed books in this work the presented ideas can be applied to any given print. However, several adaptions might be needed since the OCR on newer books is usually not based on individual training but on highly performant so-called \textit{omnifont} models. Therefore, the normalization and combination of different forms of confidence information from various engines alongside the inclusion of dictionaries might be a key ingredient.

\bibliographystyle{IEEEtran}
\bibliography{references}

\end{document}